\documentclass[runningheads]{llncs}
\usepackage{graphicx} 
\usepackage{hyperref} 
\usepackage{url}      
\usepackage{amsmath}  
\usepackage{amsfonts} 
\usepackage{xcolor} 
\usepackage{listings}
\usepackage{color}
\usepackage{listingsutf8}
\usepackage[utf8]{inputenc}

\definecolor{dkgreen}{rgb}{0,0.6,0}
\definecolor{gray}{rgb}{0.5,0.5,0.5}
\definecolor{mauve}{rgb}{0.58,0,0.82}

\begin{document}

\title{Creating a Gen-AI based Track and Trace Assistant MVP (SuperTracy) for PostNL}
\titlerunning{Gen-AI based MVP: SuperTracy}
\author{Mohammad Reshadati}
\authorrunning{M. Reshadati} 

\institute{
  Vrije Universiteit Amsterdam \and
  IT E-commerce at PostNL \\
  \email{m.s.reshadati@student.vu.nl}
}

\maketitle 
\begin{abstract}
The developments in the field of generative AI has brought a lot of opportunities for companies, for instance to improve efficiency in customer service and automating tasks. PostNL, the biggest parcel and E-commerce corporation of the Netherlands wants to use generative AI to enhance the communication around track and trace of parcels. During the internship a Minimal Viable Product (MVP) is created to showcase the value of using generative AI technologies, to enhance parcel tracking, analyzing the parcel's journey and being able to communicate about it in an easy to understand manner. The primary goal was to develop an in-house LLM-based system, reducing dependency on external platforms and establishing the feasibility of a dedicated generative AI team within the company. This multi-agent LLM based system aimed to construct parcel journey stories and identify logistical disruptions with heightened efficiency and accuracy. The research involved deploying a sophisticated AI-driven communication system, employing Retrieval-Augmented Generation (RAG) for enhanced response precision, and optimizing large language models (LLMs) tailored to domain specific tasks. 

The MVP successfully implemented a multi-agent open-source LLM system, called SuperTracy. SuperTracy is  capable of autonomously managing a broad spectrum of user inquiries and improving internal knowledge handling. Results and evaluation demonstrated  technological innovation and feasibility, notably in communication about the  track and trace of a parcel, which exceeded initial expectations. These advancements highlight the potential of AI-driven solutions in logistics, suggesting many opportunities for further refinement and broader implementation within PostNL’s operational framework. 


\begin{minipage}{\linewidth}
    \centering
    \includegraphics[width=0.5\linewidth]{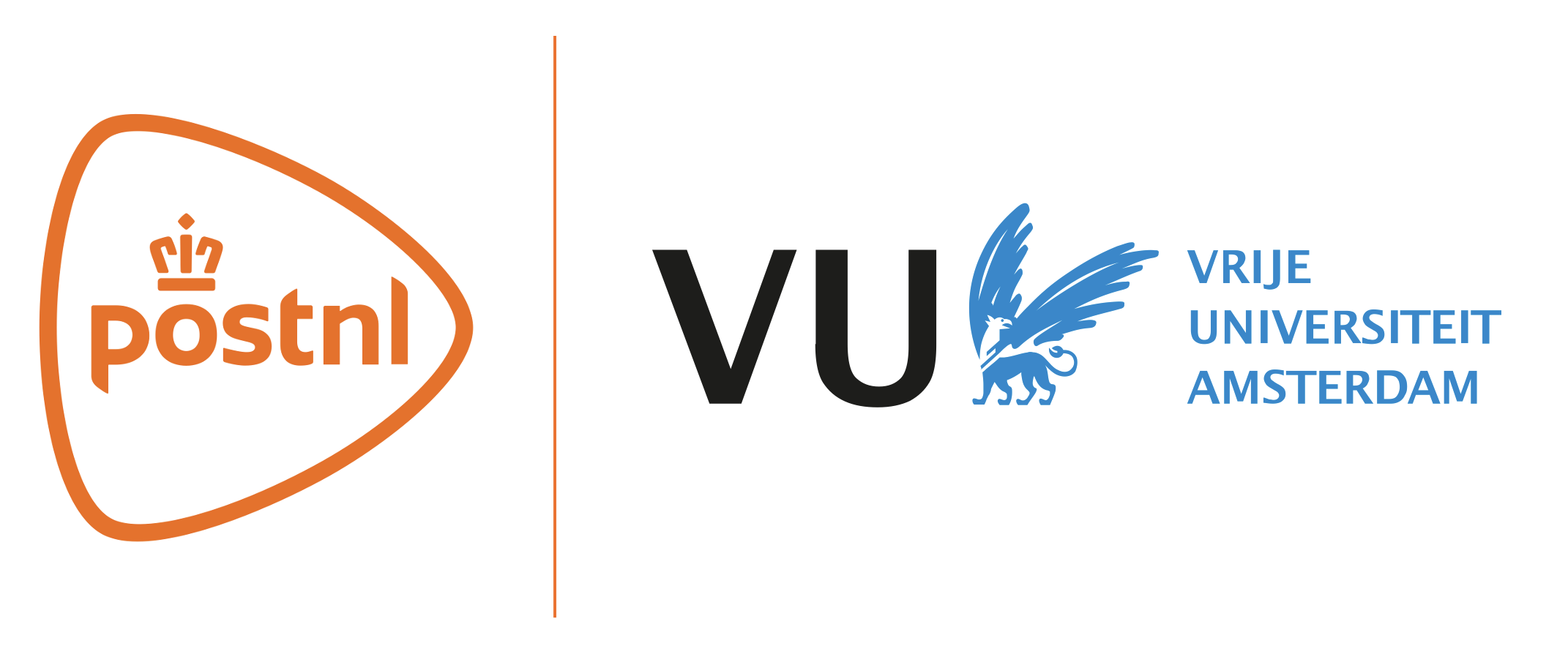}
\end{minipage}
\end{abstract}

\clearpage

\section*{Acknowledgements}

My project was greatly facilitated by my supervisors, Banno Postma from VU Amsterdam and Jochem Roth from PostNL. I thank Banno Postma for his guidance, feedback and flexibility throughout the process. I thank the Gen-AI team for welcoming me with open arms to PostNL. My special thanks to Jochem Roth for this internship opportunity, his guidance and the trust he had in me to provide the MVP and present it for the Directors. I greatly value Chris Scholtens' expertise in clarifying PostNL's logistical network and Nick Smith for providing logistics data. I express my gratitude to Jasper Bosma for his critical evaluation of the SuperTracy project, as well as to his team members Antoine Donkers, Stefan de Jong, Sebastiaan van de Koppel, Kees-Jan de Gee, Floris Groot, Robbert-Jan Joling, and Michael Street for their essential evaluations. 

\clearpage

\section{Introduction}
The developments in the field of generative AI has brought a lot of opportunities for companies. For instance, the logistics and postal sector has seen significant improvements in efficiency of customer service through the integration of AI technologies. As one of the leading postal and logistics companies in the Netherlands, PostNL has embraced these advancements to enhance the internal and external communication around track and trace of parcels. This thesis explores the development and implementation of a generative AI-based multi-agent large language model system designed to facilitate parcel-related inquiries at PostNL. By leveraging the capabilities of generative AI, this system aims to streamline communication, improve customer satisfaction and to help internal agents to understand the journey of the parcel in an easy manner.

\subsection{PostNL}
PostNL is the biggest mail, parcel and E-commerce corporation of the Netherlands, with
operations among the Benelux, Germany, Italy and the United Kingdom. It has a rich history of 225 years in The Netherlands. On an average weekday, 1.1 million parcels and 6.9 million letters are delivered throughout the Netherlands by PostNL. This huge amount is achieved by the existence of 37 sorting centers, 11.000 letter boxes, 903 Automated Parcel Lockers 5.795 retail locations and approximately 33.500 employees \cite{postnlReport}. 

The business of PostNL is high-over divided in 3 sections: parcels, mail and Cross Border Solutions (CBS). The scope of this research is parcels and E-commerce. The end-to-end process of parcel delivery can be quite complex, but very efficient at the same time. The key elements in the value chain are the following \cite{postnlReport}:
\begin{itemize}
\item \textbf{Collect} In this first step the parcels are collected from the customers. In this process, the expectations of the customer have to be matched by timely pick-up and processing.

\item \textbf{Sort} The second step is sorting and processing the parcels, based on destination and specific customer and consumer needs. An efficient sorting helps to ensure delivery to the right location on time. 

\item \textbf{Deliver} The final step is delivery, which is the moment of connection within the sender and receiver, and the final delivery of the parcel. As the delivery men are in each street in The Netherlands everyday, this gives space for additional societal services, like identifying loneliness in households and working together with charity organizations. 
\end{itemize}

The execution of these vital steps are heavily influenced by technological developments. 
The digital transformation that PostNL is going through is an important initiative to stay on top of these innovations \cite{postnlReport}. Digitalisation helps in developing the core activities to provide smart E-commerce solutions to improve its competitive position. Logistics has developed from being a pure service providing activity, to becoming a key driver of digital and societal change.  Technologies such as the Internet of Things (IoT), autonomous driving, big data and are inseparably intertwined with logistics. Recently, generative AI has emerged as another significant technology to contribute to advancements in this industry.

\subsection{Problem Statement}
The value chain of parcel delivery is described in the section above in a high-over and abstract way. However, the value chain is more complicated, depending on variables like customer and consumer preferences, the size of the parcel, the time of the year, the volume of parcels, which PostNL services are being executed and more. Throughout this process, logistic events are registered, also called 'waarnemingen' in Dutch. These logistic events represent diverse situations that can appear based on the variables. The logistic events are a code starting with a letter followed by 2 numbers. There are 400 unique logistic events. These individual events together form logistic event sequences, which describe the journey of the parcel from the moment of acceptance in the PostNL network up to delivery. There is a lot of variation possible in these sequences, resulting in hundreds to thousand different sequence possibilities. For instance, a package with a specific barcode during its journey can exhibit the following sequence of 'waarnemingen':

\[
\begin{aligned}
&\text{[}A01, A98, A95, B01, G03, V06, A04, K50, B01, A96, J01, J40,\\
&A19, J05, A19, H01, J30, B01, J17, B01, J01, J01, J40, A19, A19, J05, I01\text{]}
\end{aligned}
\]

Each code in the sequence above has a descriptive meaning and a contextual explanation. One challenge is the diverse combination of codes in sequences, where the presence or absence of a code can be due to a mistake in the logistical operational process. 

Within PostNL there are logistic business experts who can interpret these sequences and explain it to others. They know the context of the operational process behind the events. Certain combinations of logistic events in sequences can imply implicit knowledge that can only be understood and explained by these experts. They also know the full meaning of a logistic event that might not be well documented. These experts use an internal system called 'Tracy' where for the barcode of each parcel, you can look up the logistic events, along with description of the event, timestamps, where they happen (locations), source-system, and some customer and consumer information. For external communication with consumers on the status of the parcel, a small selection of these events are shared through the PostNL app notifications, through email or through the track-and-trace page on the PostNL website. 

PostNL wants to see if Gen-AI can be used to make sense of these complex logistic event sequences in a easy to understand manner. You could say an elevated version of Tracy, with the name 'SuperTracy'. Such a system could be used internally for business use-cases, or externally for consumer facing communication about the status of the parcel.

\subsection{Research Goal}
The goal of this research is to explore the potential for forming a dedicated  approach towards generative AI research, development, and engineering at PostNL, by inspiring business stakeholders on its potential. There are several sub-goals that contribute to this:

\begin{enumerate}

    \item \textbf{Exploring valuable business use-cases solved by Gen-AI:} In order to get dedication and interest of business stakeholders, it has to be shown that there are valuable use-cases that can be solved through Gen-AI. One of the use-cases is to improve the efficiency and accuracy of communication around parcel track and trace within the logistical ecosystem of PostNL through SuperTracy.
  \item \textbf{Using in-house solutions:} The MVP of SuperTracy is aimed to establish the viability of developing an entirely new in-house generative generative AI based system, eliminating reliance on external AI platforms like ChatGPT API's or Amazon Bedrock services. Doing so will decrease the costs and make the usage of Generative AI more approachable.
  \item \textbf{Creating a MVP for the business stakeholders}: Showing the value of a Gen-AI based solution to a business problem, requires a MVP. This unfolds in having a technical sound product, but also an evaluation to show the value. The evaluation has to show that SuperTracy can mimic parcel experts at least, or even enhance the understanding of parcel journeys and effectively identify and communicate issues in the logistics and delivery processes. The MVP should be suitable to show to business stakeholders to see the value.
  
\end{enumerate}

\section{Literature Study}
\subsection{Generative AI and ChatGPT}
Generative AI refers to a category of artificial intelligence algorithms that can generate new data or content that is similar to the data it was trained on. Unlike traditional AI, which typically focuses on identifying patterns and making decisions based on existing data, generative AI can create new, original content, such as text, images, music, and even code. This capability is powered by models such as Generative Adversarial Networks (GANs), Variational Autoencoders (VAEs), and transformer-based models like GPT (Generative Pre-trained Transformer). \cite{banh2023generative}. 

ChatGPT is a notable example of a transformer-based generative AI model developed by OpenAI, which is specifically designed for conversational tasks. It has attracted worldwide attention for its capability of dealing with challenging language understanding and generation tasks in the form of conversations \cite{wu2023chatgpt} \cite{teubner2023welcome}. This has also inspired businesses to make use ChatGPT AI systems, as it can have various efficiency gains. ChatGPT can automate various business tasks such as content creation, customer service, and data analysis, leading to improved productivity and cost savings. Also the model's ability to understand natural language and provide human-like responses can enhance customer engagement and satisfaction \cite{arman2023exploring}.

\subsection{Transformer Architectures and Large Language Models}
The way ChatGPT works is through sophisticated natural language processing (NLP) techniques within massive computational infrastructures that result in the fluent human like responses. The workings of this rely on neural transformer models and Large Language Models (LLMs). Transformer based models are excellent at processing longer sequences of data–like text–by using self-attention processes that enable the model to focus on different areas of the input by learning long-range dependencies in text. Self-attention is a mechanism that allows the model to learn the importance of each word in the input sequence, regardless of its position \cite{zhao2020exploring}. 

LLMs are constructed using the transformer architecture. LLMs are a type of AI model trained on massive amounts of text data to understand and generate human-like language. They have a large number of parameters, often in the billions, which enable them to capture intricate patterns in language. This scaling up has allowed LLMs to understand and generate text at a level comparable to humans \cite{teubner2023welcome}. 

So transformers help the system focus on the important parts of the input text, understanding the context and meaning. LLMs use this understanding to predict and generate the next word in a sentence, making the conversation flow naturally. 

\subsection{Open-sourced and closed-sourced LLMs}

ChatGPT is a closed-source general purpose-chatbot. The closed source LLMS are also called 'commercial' or 'Proprietary' LLMs. In general, closed-source LLM models perform well across diverse tasks, but they fail to capture in-depth domain-specific knowledge \cite{chang2024survey}. The same holds for other closed-source LLM models like other OpenAI GPT model families or Claude \cite{liu2023llm360}. The usage of LLMs for specific use-cases can sometimes seem unattainable, due to the lack of transparency, high cost and energy consumption, usage limits, and adherence to terms of service. 
The recent emergence of highly capable open-source LLMs such as LLAMA 3, T5, MADLAD, and GEMMA 2 allow researchers and practitioners at large to easily obtain, customize, and deploy LLMs in more diverse environments and for more diverse and specific use cases \cite{kukreja2024literature}. 

\subsection{Making LLMs suitable for specific tasks through fine-tuning}
There is a irresistible necessity from enterprises for fine-tuning LLMs to get them trained on proprietary domain knowledge. Fine-tuning is the process of continuing the training of an already pre-trained model on a new dataset that is typically smaller and task-specific \cite{vm2024fine}. This allows the model to adjust its weights and parameters to better fit the nuances of the new data and the specific requirements of the target task. Though there is an option to use OpenAI (open-source) models to solve most of the use-cases, there is a high demand for domain specific LLMs due to data privacy and pricing concerns as mentioned earlier. The data of an enterprise can stay on premise as the LLMs are also present on premise. In-house development ensures this. Fine-tuned LLMs provide quality and custom feel to the stakeholder and also has low latency in displaying the results.\cite{vm2024fine}

\subsection{Making LLMS suitable for specific tasks through Retrieval Augmented Generation}
When a LLM model that is not fine-tuned is used for domain specific tasks and asked to handle queries beyond its training data or current information, hallucinations can happen \cite{zhang2023siren}. As fine-tuning can be done to make LLMs suitable for specific use-cases, another approach next to or instead of fine-tuning can be to use Retrieval-Augmented Generation (RAG) architecture. RAG enhances LLMs by retrieving relevant document chunks (in real time) from external knowledge bases through semantic similarity calculation \cite{gao2023retrieval}. So RAG retrieves additional data, and augments it to the existing knowledge of the LLM based on semantic similarity. In fine-tuning, the weights of the existing parameters of the LLM get adjusted to the learned knowledge, but vectors are not added. Therefore to keep the knowledge base updated, fine-tuning would be computationally expensive.

\subsection{Enhancing the input of an LLM-based systems through prompt engineering}
The input that is given to an LLM is also important for the desired output. That input is also called a Prompt. Prompt engineering is the process of designing and refining input queries, or “prompts,” to elicit desired responses from LLMs \cite{sorensen2022information}. In the literature, prompt engineering is usually explained in two different context. Prompt engineering can apply to how a user of an LLM based system can phrase its desired task to the model the best way\cite{marvin2023prompt} \cite{bsharat2023principled}, or prompt engineering can refer to the effective way of responding of the LLM-based system, which can be determined by the developer \cite{sahoo2024systematic}. In the latter case, various prompt engineering techniques are available to guide the model effectively. Well known methods are few-shot prompting, chain-of-thought prompting and self-consistency \cite{sahoo2024systematic}. 

\subsection{Enhancing the performance of LLM-based systems through Quantization}
The performance of open-source models depend on the hardware and the available computational resources that are being used. Significant challenges can be faced when attempting to leverage the full potential of transformer models in cases where memory or computational resources are limited. Because the advancements in transformer performance are accompanied by a corresponding increase in model size and computational costs \cite{kaplan2020scaling}. Floating-point post-training quantization techniques can be used to enable the compression of transformers to face the challenges of limited computational resources. \cite{liu2023llm}. This approach enables the effective use of LLMs on hardware with constrained computational capabilities while maintaining the high quality of generative AI services. Additionally, it reduces the computational resources required for training and executing models \cite{jacob2018quantization}, resulting in cost savings.

\subsection{Logistic event prediction through sequence to sequence prediction by T5}

In the previous sections, the transformer architecture and LLMS like GPT-3 and GEMMA have been discussed. These models have demonstrated remarkable capabilities in various NLP tasks due to their ability to understand and generate human-like text. Transformer-based models are very versatile and diverse, making each of them suitable for different tasks. T5 (Text-to-Text-Transfer Transformer), developed
by Google, is a versatile language model that is trained in a ”text-to-text” framework \cite{raffel2019exploring}. The key innovation of T5 is the formulation of all tasks as text generation problems. This means that every task, including text classification, summarizing, translation, and question answering, is cast into a text-to-text format. For example, instead of training T5 to answer questions directly, it is trained to generate the complete answer given the question and relevant context. \cite{hadi2023survey}

For the prediction of future logistic events in the sequence as shown in the Problem Statement, The T5 can be used. Transformers are highly suitable for this task due to their ability to handle sequential data and capture long-range dependencies through self-attention mechanisms \cite{vaswani2017attention}. The T5 model excels in sequence prediction tasks and can be fine-tuned on specific datasets to improve accuracy \cite{raffel2020exploring}. Mathematically, the transformer architecture uses an encoder-decoder structure where both components utilize self-attention and feed-forward neural networks. The self-attention mechanism computes representations for each element in the sequence by considering the entire sequence context, enhancing the model's capability to predict the most likely sequence of logistic event codes. This mechanism can be described by the attention function, which maps a query and a set of key-value pairs to an output, computed as a weighted sum of the values, where the weights are derived from the query and corresponding key \cite{vaswani2017attention}. 

\subsection{LLMs and Multi-Agent systems}
So far various methods to make LLMs suitable for specific tasks have been discussed, such as fine-tuning and prompt engineering. As intelligent agents also focus on specific tasks, researchers have started to leverage LLMs to construct AI agents \cite{zhao2024expel}. LLMs can be employed as the brain or controller of these agents and expand their perceptual and action space. These LLM based agents can exhibit reasoning and planning abilities through earlier discussed prompt engineering techniques like Chain-of-Thought.

Based on the capabilities of the single LLM based agent, LLM-based Multi-Agents have been proposed to leverage the collective intelligence and specialized profiles and skills of multiple agents. Compared to systems using a single LLM-powered agent, multi-agent systems offer advanced capabilities by specializing LLMs into various distinct agents, each with different capabilities, and by enabling interactions among these diverse agents to simulate complex real-world environments effectively \cite{guo2024large}.

\section{The Solution}

\subsection{Data and knowledge Discovery}
To make LLMs suitable for domain specific tasks, the models have to be fine-tuned on relevant PostNL data. Several interviews have been done with experts of logistic events and data warehouse engineers to identify the existing datasets and to understand the operational process behind the data. Also the workings and background data of Tracy have been investigated. This resulted in the selection of appropriate datasets to build the solution with:

\begin{itemize}
    \item \textbf{Collo data:} The 'Collo' dataset is a well known dataset in PostNL. It contains all the barcodes of parcels, along with all the logistic events that are registered throughout the journey of the parcel, from the moment of acceptance in the network up to delivery. This is an extensive data set, with 159 columns, with each row representing an event or the current state of a parcel. 
  \item \textbf{Abbreviations:} Abbreviations are used a lot in PostNL, resulting in PostNL specific terminology which is widely used in documentation and other data. This dataset has a collection of abbreviations, along with a description and explanation. These abbreviations are also widely used throughout Collo columns and data entries. 
  \item \textbf{Waarnemingen:} This data set contains all the 400 unique 'waarnemingen' or logistic events, along with a description of what each mean. Each Waarneming code of a parcel forms a row in Collo.  Additionally, each code falls into either the internal or external category.  The internal category indicates that this event is only visible for PostNL employees through Tracy for example. The external category indicates that the logistic event is being shared externally with customers. For example through the PostNL app or email. 
  \item \textbf{Location data:} The location master data set contains all the locations that PostNL does business on. Examples are warehouses, sorting centres, distributions centres, hubs and more. 
  
\end{itemize}
For this project access to Tracy was provided. Tracy is the web-application containing all the data on parcels in real-time. Tracy ingests Collo data. Tracy has been used throughout the process to lookup some barcodes for tests. The combination of the mentioned data sources allowed for a meaningful interpretation of the main dataset, Collo.

\subsection{Data Preparation}
The datasets procured from the data warehouse were initially in a raw format, necessitating extensive data preparation prior to further analysis. The initial phase involved Exploratory Data Analysis (EDA), through which a comprehensive understanding of the dataset was developed. This was followed by the data preparation phase, outlined as follows:

\begin{itemize}
    \item \textbf{Data Cleaning:} This process addressed issues such as missing values and duplication of data points to ensure the integrity of the dataset.
    
    \item \textbf{Statistical Analysis:} Statistical summaries were performed to examine the distribution, mean, median, mode, and variance of the data. This analysis was critical for understanding the underlying patterns and anomalies within the dataset.
    
    \item \textbf{Data Transformation:}
    \begin{itemize}
        \item \textbf{Language Standardization:} Translated 132 columns of logistic parcel data from Dutch to English to establish a uniform language baseline essential for subsequent project implementation.
        \item \textbf{Normalization and Cleansing:} This step involved both normalizing and cleansing the textual data to enhance its suitability for analysis. Normalization tasks included case conversion to minimize case sensitivity issues and tokenization to structure the text into usable segments. Concurrently, the data was cleansed by removing punctuation and special characters to prevent potential data processing errors. These processes together ensured that the textual data was not only uniform but also clean and optimized for subsequent analytical tasks.
    \end{itemize}
    
    \item \textbf{Data Splitting:} The dataset was segmented into training, validation, and test sets to support the development of robust predictive models.
\end{itemize}

These preparatory steps were instrumental in ensuring that the data was aptly conditioned for the sophisticated analyses and modeling that followed.

\subsection{Model design}

\subsubsection{Expected model output}
The first step of design thinking \cite{hacker1998design} is to understand the problem, and then have clear what the concept of the solution can be. So to understand what kind of output is expected. This was achieved by discussions with domain experts, creating a shared agreement on the possible expected outcomes of the final system. The agreed goal of the system is to simulate the comprehension and narration abilities of a logistic domain experts at PostNL, who answers in a user friendly and helpful manner. 

\subsubsection{Overall Model design}

To achieve the goals stated in section 1.3, a solution is proposed building upon the findings of the literature study in section 2. The final solution is SuperTracy, a multi-agent LLM based system, leveraging open-source LLM models like GEMMA 2 and LLAMA 3 to make sure the company data is on premise and safe. Prompt engineering has been applied to direct the agents to a specific behaviour and output style. This solution is further fine-tuned on relevant datasets to make it suitable for specific use-cases. The system is build upon the RAG-architecture, such that hallucination is prevented and realtime data can be used. Also the solution is quantized, to make sure it can be run on modest hardware. The final solution contains different models and modules like logistic event prediction and translation to different languages for extra features.

\subsubsection{language detection and translation}
LLMs already have the capability to 'speak' in different langauges, but the performance is limited for some languages \cite{zhao2024llama}. In order for an LLM based system to function, the instructions through prompt engineering are necessary. These prompt engineering templates need to be in the same language as the destination language. As the templates are designed by the developer, they are predefined. For this solution, the prompt templates are available in English and Dutch. But if another language is desired, the templates need to be translated. To do this first the language has to be detected from the user input, and then translation can happen. The CLD3 \cite{ooms2024cld3} neural network detects the language of the user input. As mentioned earlier in the literature review, T5 models are suitable for translation tasks \cite{raffel2019exploring}. The MADLAD model, which is a variation of the T5 model, is employed for text-to-text translation of the templates.

\subsubsection{Logistic event prediction}

On the moment of data retrieval, not all the 'waarnemingen' codes for parcels were complete, as some parcels were still in progress and their final code had not been generated yet. To explore LLMs for sequence-to-sequence problems, the T5 model is used to predict the most likely sequence of 'waarnemingen' codes to make up for the absence of the end state of the journey. By using this strategy, we can predict the future states of parcels. If a 'waarnemingen' code indicates a problem, we can anticipate which combinations of 'waarnemingen' codes will occur so that pro-active communication can take place. This allows us to complete the entire sequence of 'waarnemingen' codes and estimate the final stage of the parcels' journeys. In the generated output it will be stated what the prediction is. 

Given an input sequence of 'waarnemingen' codes \( X = [x_1, x_2, \ldots, x_n] \), where each \( x_i \) represents a specific code (e.g., "A01", "A98"), the goal is to predict the most likely next code in the sequence. Each 'waarnemingen' code is converted into a dense vector representation through an embedding layer. The self-attention mechanism computes the attention scores for each pair of codes in the sequence to capture their relationships:

\begin{equation}
\text{Attention}(\mathbf{Q}, \mathbf{K}, \mathbf{V}) = \text{softmax}\left(\frac{\mathbf{Q} \mathbf{K}^T}{\sqrt{d_k}}\right) \mathbf{V}
\end{equation}

Multiple self-attention mechanisms (heads) capture different aspects of the relationships:

\begin{equation}
\text{MultiHead}(\mathbf{Q}, \mathbf{K}, \mathbf{V}) = [\text{head}_1; \text{head}_2; \ldots; \text{head}_h] \mathbf{W}_O
\end{equation}

The output of the multi-head attention is passed through a position-wise feed-forward neural network to introduce non-linearity:

\begin{equation}
\text{FFN}(\mathbf{x}) = \max(0, \mathbf{x} \mathbf{W}_1 + \mathbf{b}_1) \mathbf{W}_2 + \mathbf{b}_2
\end{equation}

Finally, the transformer's decoder predicts the next 'waarnemingen' code by applying a softmax layer over the vocabulary of possible codes:

\begin{equation}
P(x_{n+1} | x_1, x_2, \ldots, x_n) = \text{softmax}(\mathbf{h}_n \mathbf{W}_O + \mathbf{b}_O)
\end{equation}

Using these mathematical formulations \cite{vaswani2017attention}, the T5 model processes the input sequence of 'waarnemingen' codes, capturing dependencies and predicting the most likely subsequent codes, thereby allowing us to anticipate future states and resolve issues in parcel transportation \cite{devlin2019bert}.

\subsubsection{Fine-tuning LLAMA}
LLAMA focuses on the main expected task of the solution: describing the journey of a specific barcode based on its associated data dependencies. In this step the advanced capabilities of the latest open-source LLM, specifically the LLAMA 3 model developed by Meta \cite{MetaLlama} are leveraged. The objective is to fine-tune this model using the prepared training datasets as described in 3.1.
The training dataset was constructed following the Alpaca-style methodology\cite{taori2023alpaca}, where synthetic data is generated to mimic real-world scenarios. This methodology involves structuring data in a specific format that consists of an instruction, contextual input, and the model's expected response. For example, the training set schema  looks like this:

\begin{lstlisting}[language=Python]

alpaca_training_schema = """Below is an instruction that describes a task, paired with an input that provides further context. Write a response that appropriately completes the request.
### Instruction:
{Describe the key events in the package's journey from sender to receiver, focusing on crucial moments.}
### Input:
{Tracking details indicating times, locations, and statuses.
}
### Response:
{A concise narrative summarizing the packages journey, highlighting important transitions and updates.
}"""
\end{lstlisting}

This structured approach allows us to create a high-quality, domain-specific training dataset without extensive manual annotation, typically required. By integrating these training datasets, the LLAMA model not only retains its comprehensive linguistic comprehension but also gains specialized knowledge crucial for interpreting parcel statuses and journeys effectively.

Fine-tuning the LLAMA 3 model with our enriched dataset is executed using the Hugging Face Supervised Fine-tuning Trainer\cite{huggingface2024sft}. The optimization objective is to minimize the loss function over the training data, typically using cross-entropy loss for language models:

\begin{equation}
\mathcal{L}(\theta) = \frac{1}{N} \sum_{i=1}^{N} \mathcal{L}(\hat{y}_i, y_i)
\end{equation}

where \( \theta \) denotes the model parameters, \( N \) is the number of training samples, \( \hat{y}_i \) and \( y_i \) represent the predicted and true labels, respectively, and \( \mathcal{L} \) is the loss function \cite{vaswani2017attention}. The training process involves techniques such as gradient accumulation and mixed-precision training to manage memory consumption and accelerate computation. Gradient accumulation effectively increases the batch size without requiring additional memory, which is crucial for handling large models. By carefully adjusting parameters and leveraging these advanced training techniques, we ensure that the fine-tuned model achieves optimal performance.

\begin{figure}
    \centering
    \includegraphics[width=1\linewidth]{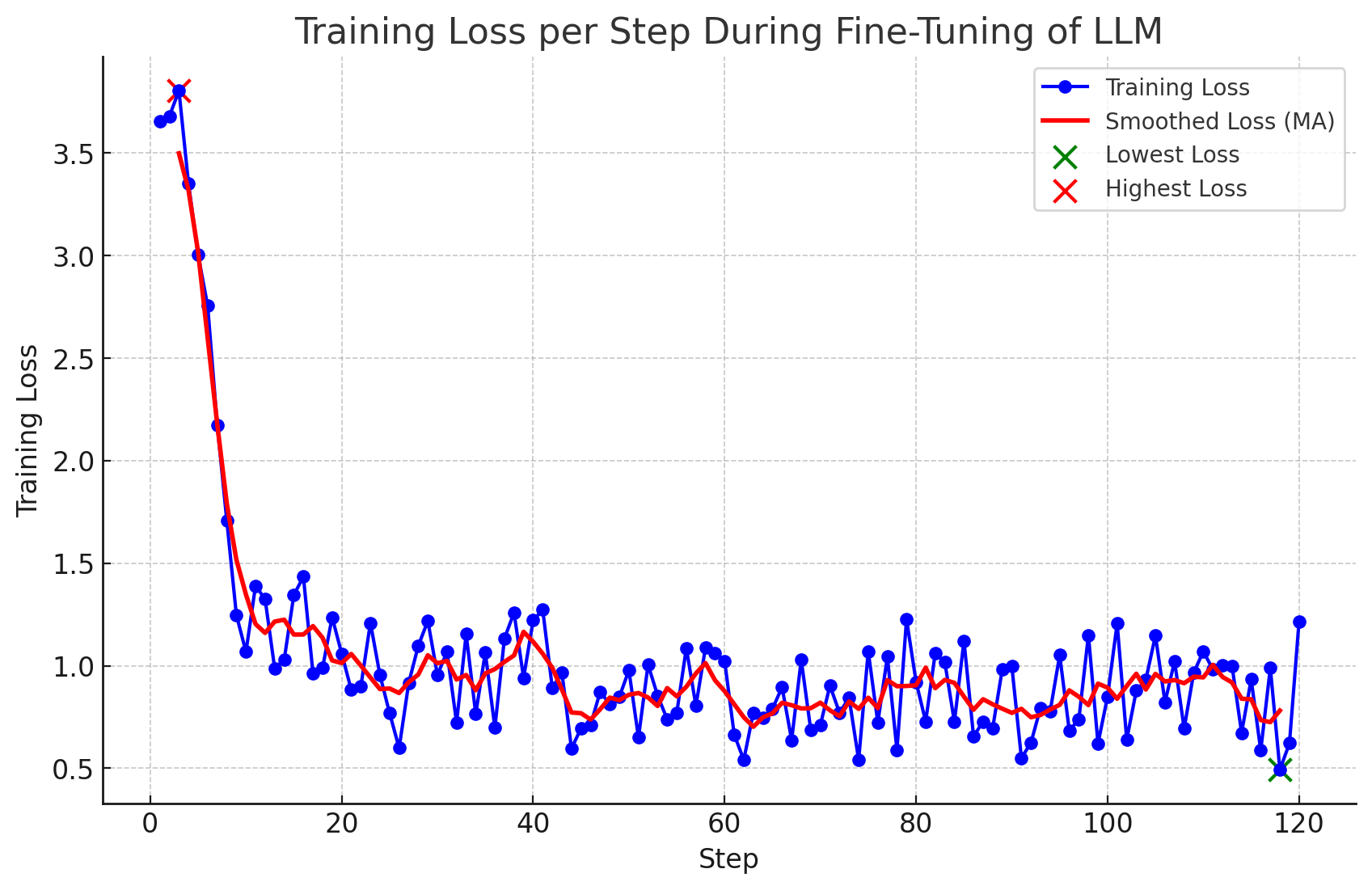}
    \caption{Fine-Tuning LLAMA3}
    \label{fig:enter-label}
\end{figure}

The plot in figure 1 delineates the training loss trajectory of a Large Language Model (LLAMA3) \cite{MetaLlama} during a fine-tuning phase conducted over a single epoch, comprising 120 steps. The model configuration involves 83,886,080 trainable parameters. Key training parameters include a per-device train batch size of 2, a gradient accumulation strategy across 4 steps, and an initial learning rate of \( 2 \times 10^{-4} \), optimized using an 8-bit AdamW optimizer with a linear learning rate scheduler. The primary plot line, marked in blue, represents the raw training loss recorded at each step, reflecting the model's immediate response to batch-level optimizations. Overlaying this, a red trend line—calculated as a moving average—smoothes out fluctuations to highlight broader trends in model performance and stability. Noteworthy are the annotations at the lowest and highest points of loss, which pinpoint critical moments in the training process where the model achieved optimal learning and where it may have struggled, respectively.

\subsection{Optimizing the performance of the model}

To optimize the model's performance during fine-tuning, quantization is being applied. Quantization  reduces the memory footprint and computational requirements of neural networks. Quantization involves reducing the precision of the model parameters, typically from 32-bit floating-point (FP32) to lower bit-width representations such as 16-bit floating-point (FP16), 8-bit integers (INT8), or even 4-bit integers (INT4). Mathematically, this can be expressed as:

\begin{equation}
\tilde{W} = \text{round}\left(\frac{W - \min(W)}{\Delta}\right) \cdot \Delta + \min(W)
\end{equation}

where \( \tilde{W} \) represents the quantized weights and \( \Delta \) is the quantization step size, defined as \( \Delta = \frac{\max(W) - \min(W)}{2^b - 1} \), with \( b \) being the number of bits used in the quantization \cite{jacob2018quantization}. This transformation maps the original weight values into a discrete set of levels, significantly reducing the number of bits required to store each weight. This process is crucial for deploying large models on resource-constrained devices, allowing efficient storage and faster computation without significantly compromising model performance.

Another employed technique  is Low-Rank Adaptation (LoRA), which enhances the efficiency of large language models by updating only a small subset of model parameters. LoRA uses low-rank matrices to approximate the updates to the weight matrices in the model, thereby reducing the computational burden. Formally, let \( W \in \mathbb{R}^{d \times k} \) be the original weight matrix. $d \times k$ are the dimensions of the original weight matrix, where $d$ is the number of output features and $k$ is the number of input features. LoRA approximates \( W \) as:

\begin{equation}
    W \approx W_0 + \Delta W
    \label{eq:W_approx}
\end{equation}

Where the change in \( W \), denoted \( \Delta W \), is given by:
\begin{equation}
    \Delta W = A B
    \label{eq:deltaW}
\end{equation}

where \( W_0 \) is the pre-trained weight matrix, and \( A \in \mathbb{R}^{d \times r} \) and \( B \in \mathbb{R}^{r \times k} \) are low-rank matrices with \( r \ll \min(d, k) \). This decomposition reduces the number of parameters from \( d \times k \) to \( r(d + k) \), resulting in significant computational savings. The optimization problem during fine-tuning focuses on learning the matrices \( A \) and \( B \), rather than the full weight matrix \( W \) \cite{hu2021lora}.

Quantization facilitates inference, particularly on devices with limited resources. Quantization during the training process ensures that the model is low-precision and robust. To ensure numerical stability,  quantization should be configured in advance and gradients should be computed in reverse order with floating-point precision. By decreasing computational and memory consumption, quantization and LoRA facilitate the utilization of LLAMA 3 by quite modest hardware only.

\subsection{Architectural design}

\subsubsection{The multi-agent setup}
The main idea of having this multi-agent LLM-based system revolves around domain-specific knowledge agents characterized by role-based agents and system template prompt engineering. This approach determines the main goals of each specific agent and constrains their behavior based on specified requirements and the state of the environment. In this context, the environment is the SuperTracy interactive system, encompassing user interactions and the contextual states of the conversation. There are three agents defined:
\begin{enumerate}
    \item \textbf{Reception Agent}: The Reception Agent handles basic communication. It introduces itself to the user and provides guidance on using the system, including instructions to provide the parcel's barcode. This agent ensures a smooth initial interaction and prepares the user for further engagement with the system.
    \item \textbf{Parcel Agent}: The Parcel Agent is responsible for analyzing parcel data and generating detailed narratives of parcel journeys. These narratives can range from detailed reports to short, coherent answers, depending on user needs. This agent includes specialized sub-models, such as a predictive model for forecasting the future status of parcels. By customizing its responses, the Parcel Agent enhances the user's ability to track and understand parcel movements comprehensively.
    \item \textbf{Knowledge Expert Agent}: The Knowledge Expert Agent specializes in answering user questions related to internal PostNL concepts and domain-specific terms. This agent can handle queries ranging from simple explanations to complex scenarios. Its knowledge base is derived from PostNL's internal documents and general logistics knowledge. The agent's ability to reason and provide contextually accurate answers makes it a valuable resource for users seeking detailed information on PostNL operations.
\end{enumerate}

The Reception agent and the Parcel agent contribute to the expected outcome of SuperTracy, that is to communicate on the parcel's track and trace journey. The Knowledge Expert agent is an additional bonus feature. The idea emerged beyond the scope of the main research question and the main requirements of the project which was only limited by the parcels track and trace. The knowledge expert agent is developed using the same approach as the Parcel agents, it is trained on internal PostNL documents. This enables it to answer and explain questions related to internal PostNL knowledge, providing reasoning based on this knowledge. It can handle simple queries, such as explanations of PostNL-specific abbreviations or business and technical terms, as well as more sophisticated questions, such as offering advice on logistical scenarios with certain problems. In all cases, the model answers questions using its acquired knowledge and foundational reasoning capabilities. You can see examples of simple and complex queries in Appendix A.

The agents together generate parcel journeys that are comprehensible due to the use of plain language instead of a sequence of logistic events. Through prompt engineering, the model exhibits both efficiency and adaptability, generating concise stories of parcel journeys that incorporate contextual information. 
\subsubsection{Prompt engineering}
The developer of a LLM-based system can instruct the behaviour of the system through prompt engineering. For this solution Chain-of-Thought prompting and Few-Shot prompting are used \cite{sahoo2024systematic}. Chain-of-Thought prompting is a technique to prompt LLMs in a way that facilitates coherent and step-by-step reasoning processes. Few-shot prompting provides models with a few input-output examples to induce an understanding of a given task.

Within prompt engineering a template refers to a natural language scaffolding filled in with raw data, resulting in a prompt \cite{sorensen2022information}. Throughout the solution, 4 different types of templates have been identified.

\begin{enumerate}
    \item \textbf{Template for cognitive behaviour of the agents}: In this template the agent is told how to behave. This differs for each agent. Here you can see the prompt template for the Reception agent: 
\begin{lstlisting}[language=Python]
    def receptionagent_context_en(): 
    return(
        """You are SuperTracy, an AI agent helping PostNL customers with their parcel tracking needs. You provide detailed information about the journey of parcels using a given barcode. Your primary function is to guide users to provide the barcode of their parcel, and then use that information to fetch and relay tracking details. You respond in a helpful and professional manner, always prompting users for the barcode if it hasnt been provided, and handling errors or questions about parcel tracking gracefully. If the barcode provided does not return any information or is invalid, you instruct the user on how to find a valid barcode or suggest alternative solutions. When you do not know the answer to a user's question, respond with, I'm sorry, I don't have information on that topic. Please provide a barcode if you need tracking details. Your interactions are designed to be clear, concise, and focused on parcel tracking to enhance customer service efficiency."""
    )
\end{lstlisting}
    
    \item \textbf{Template for instructions for each agents}: The instruction templates tell agents what to do. In this template, variables can be used which refer to other prompt templates. In the example below, you see the variable context-str which refers to the template of the parcel report. Using that template, the agent can act based on the instructions.
    
\begin{lstlisting}
    def parcelagent_context_en(context_str):
    return (
        """You are a PostNL customer service AI. You have been provided with a comprehensive overview of the journey of a parcel. This overview includes timelines, detailed event descriptions, and insights into the parcel's handling and route."
        "Here are the relevant details for the context:"
        "{context_str}"
        "Instruction: give a concise and short story about a parcel's journey from shipment to potential delivery. Highlight key events and movements between sorting centers, important timelines, and the logistics of parcel handling. Incorporate predictions regarding the future states of the parcel.

 """
    )
\end{lstlisting}

    \item \textbf{Template for creating a parcel report}: This template is based on Chain-of-Thought and introduces a step-by-step reasoning process for the agent on what to do with the input prompt. There are various steps which cant all be included here. But the general reasoning process for the agent is to extract the barcode from the input, and gather the related information from the provided data set. The key data is structured in a template, and passed to the parcel agent to generate a response with. 
    \item \textbf{Template for generating the output}: In this prompt template variables like the memory of the agent, the context window, the chat mode and optional follow up questions to ask are determined. Also the temperature of the agent can be defined, which is a parameter that controls the randomness of the model's output, affecting how predictable or creative the generated text is. Based on this template, the agent will generate the response.

\end{enumerate}

\subsubsection{RAG architecture }
To address challenges and drawbacks of the fine-tuned model like hallucinations and performance issues, we leverage the power of RAG architecture and a vectorized database. RAG architecture combines retrieval and generative capabilities to enhance LLM responses. The architecture has two components: the retriever and the generator. The retriever fetches relevant documents or data segments from a large corpus, while the generator creates responses based on the retrieved information \cite{karpukhin2020dense}. Vector databases store data in vector format for efficient similarity searches. Embedding models convert textual data into high-dimensional vectors where semantically similar texts are closer together \cite{lewis2020retrieval}. Hence, the similarity between a document \( \mathbf{d} \) and a query \( \mathbf{q} \) is computed using cosine similarity:

\begin{equation}
\text{sim}(\mathbf{q}, \mathbf{d}) = \frac{\mathbf{q} \cdot \mathbf{d}}{\|\mathbf{q}\| \|\mathbf{d}\|}
\end{equation}

Using “mxbai-embed-large” as embedding model, the mxbai-embed-large model, a transformer-based architecture, generates high-dimensional, semantically rich embeddings. It uses multi-head self-attention mechanisms, weighing the importance of different words in a sentence dynamically. The attention score \( \alpha_{ij} \) is calculated as:

\begin{equation}
\alpha_{ij} = \frac{\exp(e_{ij})}{\sum_{k=1}^{n} \exp(e_{ik})}
\end{equation}

where \( e_{ij} = \mathbf{q}_i \cdot \mathbf{k}_j \). The model passes input text through multiple layers of attention mechanisms and feed-forward neural networks, producing a dense vector that captures the text's semantic meaning. This embedding is used for retrieval in RAG architectures. Mxbai-embed-large achieves state-of-the-art performance in NLP tasks, making it a reliable choice for embedding-based retrieval in RAG architectures \cite{lee2024open}. Optimizing the retriever, generator, and vector database is contributes to scalability. Verifying and validating generated responses are crucial to mitigate hallucinations \cite{li2023angle}.

In order to input the parcel data into the implemented RAG architecture, an ETL pipeline is developed to Extract, Transform, and Load the parcels data. This pipeline consists of the following three steps:

\begin{itemize}
    \item \textbf{Extract}: Data is retrieved from the earlier mentioned sources (3.1), the main one being Collo.
    \item \textbf{Transform}: Data is cleaned, normalized, and structured; for example, date formats are standardized, and missing values are handled.
    \item \textbf{Load}: Transformed data is loaded into a vectorized database, ready for querying and analysis by LLMs.
\end{itemize}

The integration of RAG architecture and the developed ETL pipeline enhances the precision and contextual richness of the responses of SuperTracy.

\subsection{Overall System Architecture Overview}

In the diagram below (Figure 2) you can see the architecture of the final system, where all the components come together that form SuperTracy. The architecture of the system is designed to ensure integration and effective collaboration among the agents and models. Key components include:

\begin{figure}
    \centering
    \includegraphics[width=1\linewidth]{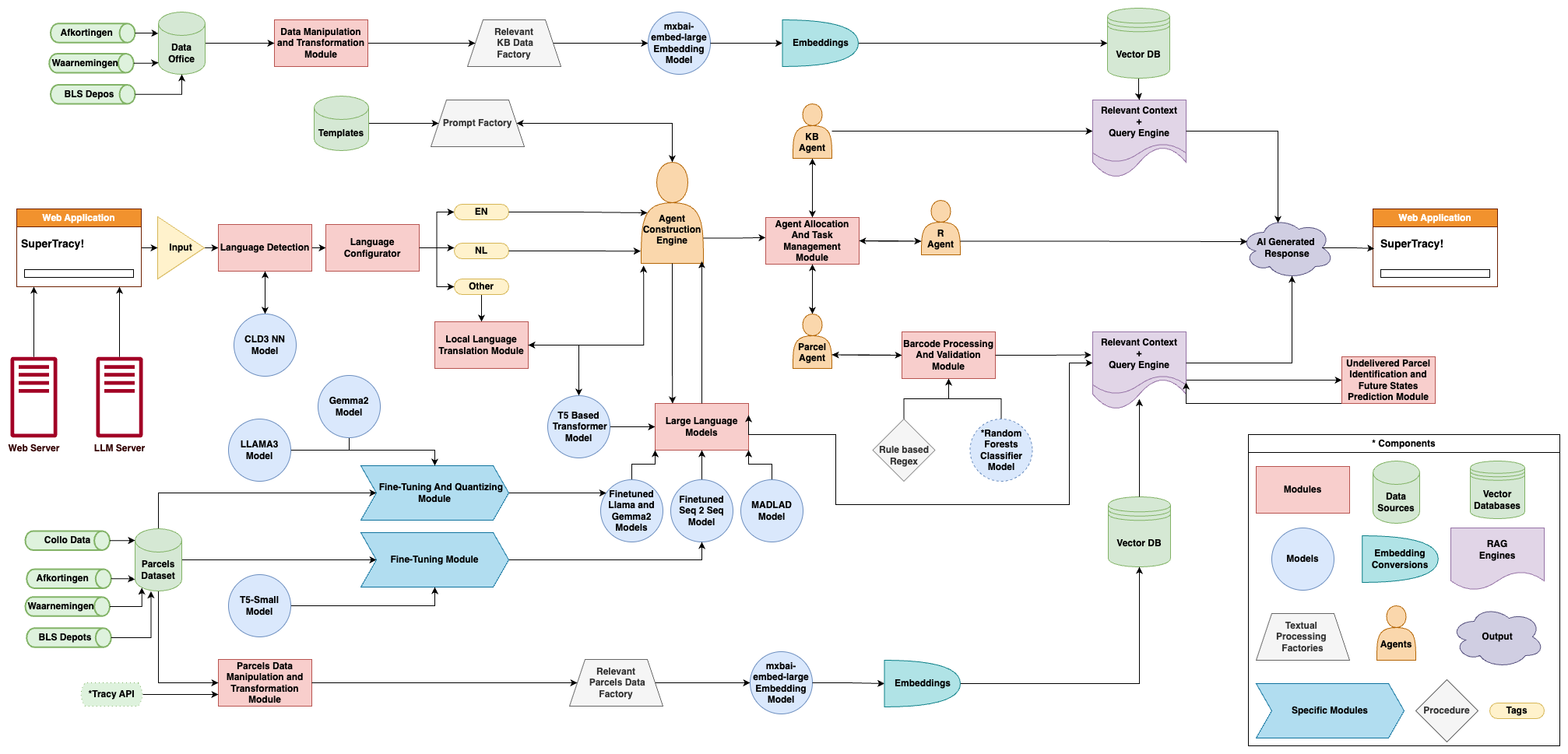}
    \caption{\href{https://drive.google.com/file/d/1QBwRVuOrS9tdSuwtSfhn0g1A_H8fq8a8/view?usp=sharing}{SuperTracy System Architecture}}
    \label{fig:enter-label}
    \href{https://drive.google.com/file/d/1QBwRVuOrS9tdSuwtSfhn0g1A_H8fq8a8/view?usp=sharing}{*\textit{To view a higher resolution version of the SuperTracy architecture, please click here.}
}
\end{figure}

\begin{itemize}
    \item \textbf{LLAMA3 and GEMMA2}: LLMs serving as the backbone for different agents, providing advanced language understanding and generation capabilities \cite{MetaLlama,gemma2024improving}.
    \item \textbf{CLD3}: A neural network model for detecting the language of user queries, ensuring accurate processing of inputs \cite{ooms2024cld3}.
    \item \textbf{MADLAD}: A text-to-text translation model facilitating multilingual support \cite{kudugunta2023madlad}.
    \item \textbf{mxbai-embed-large}: An embedding model used in the RAG architecture to enhance retrieval and augmentation processes \cite{lee2024open}.
    \item \textbf{Random Forests Classifier and Regex Pattern Matching}: Models and methods for verifying the correctness of barcodes, enhancing the system's reliability.
    \item \textbf{T5 model}: This model is used for logistic event prediction of the parcel. 
    \item \textbf{Prompt Factory}: Prompt factory is the module where prompt engineering takes place and various prompt templates are created and managed.
\end{itemize}

The integration of these components forms a cohesive system capable of addressing various user needs through the Reception, Parcel, and Knowledge Expert Agents. Each agent operates within its specialized domain, leveraging the strengths of the underlying models to provide accurate, relevant, and timely responses.

\subsubsection{The final product and User-Interface}
This final stage required integrating all components and sub modules into a unified architecture, involving the following key steps:

\begin{itemize}
    \item \textbf{System Integration}: Ensuring all components, including the fine-tuned LLMs, the RAG architecture, and the agents work together seamlessly was a challenging software engineering task. This included writing code, debugging, and optimizing system performance to handle user queries efficiently.
    \item \textbf{Web Platform Development}: Creating a user-friendly web interface that allows users to interact with the system. This platform serves as the front end, providing an intuitive and accessible means for users to query the knowledge expert agent and receive responses. The web-interface is visible in the figures in Appendix A.
\end{itemize}

The successful execution of these procedures has led to the development of a comprehensive system that combines multiple LLMs into a fully operational and user-interactive system. This system not only provides answers to basic questions about PostNL-specific terminology but also offers advanced guidance on logistical situations, showcasing the advanced reasoning abilities of the LLMs.

\section{Evaluation and Discussion}

\subsection{Technical Evaluation of the model}
SuperTracy is a MVP which hasn't been deployed yet. Therefore, technical performance cannot be formally measured. Below are a few important implementation goals, that made the the finalization of  this MVP possible. These implementation goals build further on the three research goals mentioned in section 1.3.
\begin{itemize}
    \item \textbf{Lightweight Deployment}: The system operates effectively on local machines with modest hardware. This has been achieved by using quantization and open source LLM models. For the deployment of the MVP of Supertracy, a MacBook Pro with chip M1 Max and 64 GB memory has been used. The reaction time of the system was always less than 2 second. Videos are taken of the performance.
    \item \textbf{Complete Local Integration}: All necessary modules and models used for SuperTracy are integrated and run entirely on the local system. This means that external API’s like Open-AI or Bedrock API’s are not used, which can be quite expensive on a large scale. Open source models like GEMMA 2 and LLAMA 3 have been used instead, which also ensure the company data stays on premise. 
    \item \textbf{Integration with RAG Architecture}: The integration of agents with RAG architecture, allows agents to work individually or together to perform diverse and complex tasks. Using RAG has made it possible to specialize the LLMs for specific tasks, therefore enabling the  business use-case. Through RAG, agents are able to utilize embedded documents to enhance their knowledge base.  RAG also enables the system to recall its reasoning and used resources in contrast to the closed-source external LLMs which act as a black box.
\end{itemize}

\subsection{Human Evaluation of the generated output by SuperTracy}
The research goals stated in section 1.3 state that a LLM-based system can be build in-house, and solve the business use-case of mimicking the role of a logistical expert, making sense of logistical events of a barcode and being able to communicate about it. 

Evaluation of LLM models are challenging, as the output is diverse each time and  can be evaluated against a variety of different metrics, like fluency, accuracy, trustworthiness, presence of bias, factuality, multilingual tasks, reasoning and more \cite{chang2024survey}. Also the task of the LLM can be diverse, for example summarizing, translating, question and answering or casual conversing. Therefore the first question is 'what' to evaluate. For the case of SuperTracy, it has been decided together with the business experts, to evaluate on the factual and relevant information given in the parcel story. Aspects like fluency or multilingual performance and the performance of Knowledge Expert agent and recipient agent are left out of scope. 

The second question is 'how' to evaluate. The two common evaluation methods are automatic evaluation and human evaluation \cite{chang2024survey}. For the evaluation of SuperTracy, Human evaluation is chosen, as  available automated evaluation techniques or benchmarks are not suitable for evaluating factuality of the enterprise specific generated parcel stories \cite{chang2024survey}. When choosing human evaluation methods of LLMs, attention has to be paid to various crucial factors to guarantee the dependability and precision of assessments \cite{singhal2023large}. Important criteria are the number of evaluators being around 9 \cite{belz2006comparing}, evaluation criteria ('what' to evaluate), and evaluator’s expertise level. 

\subsubsection{Experimental Setup of Human Evaluation}
A panel of eight logistical domain experts of the supply chain team have been asked to critically review the output of SuperTracy based on the factual correctness of the generated parcel stories. 

A sample of 100 parcel barcodes have been selected, with the minimal requirements provided by the logistics expert to ensure that they represented complex logistical scenarios likely to challenge the system's capabilities. These requirements have been provided by mentioning which 'waarneming' codes indicate an unhappy journey flow. The sample of these barcodes containing the selected events have been given to the model to generate a parcel story for. 

These experts were asked to assess the factual correctness and relevance of the stories on a scale ranged from 1, indicating a very low level of correctness and relevance, to 5, representing an exemplary level of performance in the parcel story generation. Participants were asked to provide an explanation for a score lower than 3. In that way feedback and insight into lower scores have been achieved. It was decided together with the business experts, that a score of 3 was good enough and anything below indicated a low quality of the generated stories.  

\subsection{Results of the evaluation of Domain Experts}
\subsubsection{Quantitative Results}
The results of the given scores by the domain experts on the generated parcel stories by SuperTracy are shown in the figure below. The AI generated outputs are generally well-received by the logistics domain experts.  The most common scores are 3 and 4, with a median score of 4. 75\% of the generated parcel stories got the score of 3 or more. This reflects that most domain experts rated SuperTracy's performance favorably. Comparing these results to the  expected score of 3 or higher set by the domain experts, indicates that the performance of SuperTracy has been good enough to prove that it can solve the problem statement. 
\begin{figure}
    \centering
    \includegraphics[width=1\linewidth]{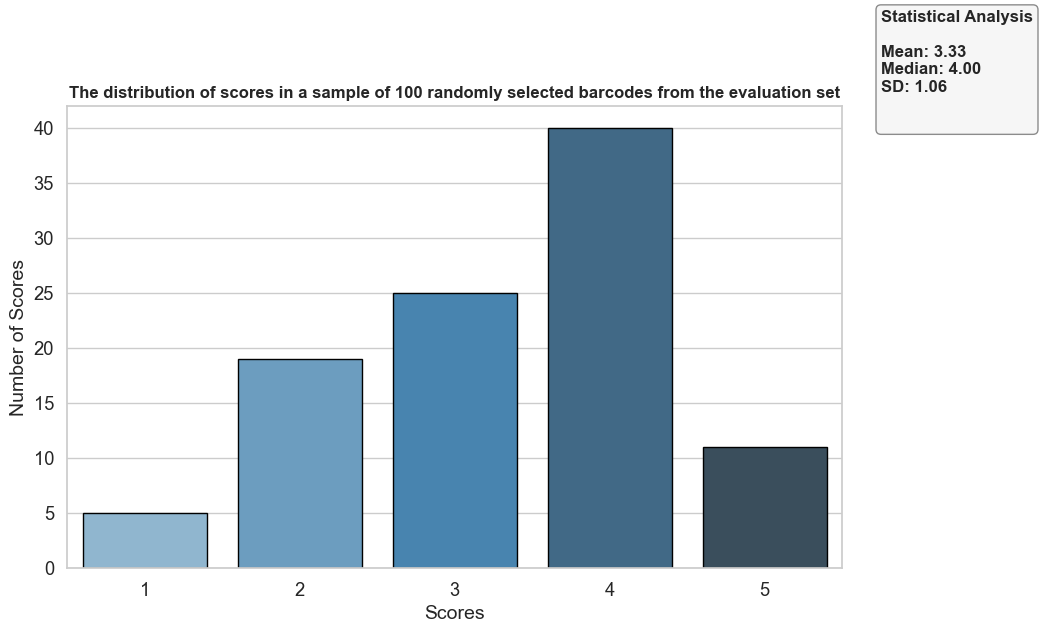}
    \caption{Result of the Human Evaluation}
    \label{fig:enter-label}
\end{figure}

\subsubsection{Qualitative Feedback}
For the scores of lower than 3, the domain experts have provided reasoning for their choice. The main gist of the gathered feedback from the open answers is about incorrect generalizations or incorrect assumptions of a few event codes. These are summarized in the following points:
\begin{itemize}
    \item \textbf{Incorrect assumptions and generalizations of some logistic events:} For example in the logistic events 'ETA was updated' happens quite a few times at the end of a logistical sequence. This is interpreted as a delay by the model, but this is not necessarily a delay. Also according to the standard process of evening distribution 'Avonddistributie' the route gets rescheduled, as they are in a planned network. Or in the morning planners can shift parcels from routes to improve the total planning. This has nothing to do with 'delay' or 'unforeseen circumstances' as SuperTracy calls it in the parcel stories. 
    
    \item \textbf{Some steps are default in the process and are not interesting to show}: Sometimes it is possible to change something. This is present in the logistic event. For example, the event 'Changing ETA is possible'. But this does not mean that it has to be anticipated on, as its an automatic default released event. This doesn't have to be mentioned in the parcel story. Only the actual made changes need to be mentioned. An example is that before the first sorting there is always a notification that 'it is not possible to change the date or time'. This is interpreted as a problem, but actually nothing went wrong. The information that is not valuable to share, should be distinguished upfront to make the outcome more clear to the recipient. 
    \item \textbf{Data quality issues in location data}: The location data had some issues. Most of the locations were identified as sorting centres, but not all locations are sorting centers. There is a difference between sorting, distribution and retail locations. Not distinguishing between these different locations which execute different business processes, makes the parcel story less accurate.
    
\end{itemize}

\subsection{Conclusion}
The goal for this project was to create a MVP which can act as a show-case of the value that generative AI can bring for PostNL. Reflecting on the goals set at the beginning of the research in section 1.3, we can conclude that the goals have been achieved.
The main goals were to explore business use-cases that can be solved by Generative AI and creating a MVP to showcase the value to the business. Throughout the demo and the evaluation of SuperTracy by domain experts, a lot of positive reactions are received. The use-case of SuperTracy has inspired the supply chain team to consider using such a system to improve their workflow, and for the business stakeholders to further explore possibilities of refining for deployment. The demo of SuperTracy has also inspired the stakeholders to realize the value of leveraging generative AI techniques, by brainstorming on various new use-cases during the demo. Also the goal to use in-house solutions when designing and creating the system was achieved, by using open source LLM models and using local computing power for running the system. To further prove the value of the use-case, the  Human Evaluation confirms that SuperTracy is a successful MVP that can mimic the track and tracing capabilities at PostNL by effectively creating a story of the parcel's journey, receiving a score of higher than 3 in 75\% of the cases. Interestingly, the feedback provided by the subject matter experts, were mostly about the quality of the input data, and not the LLM-based model itself. At the same time, PostNL is working on strengthening their data fundament, which will result in better data quality, which forms the base of a good performing LLM system. 

All together, the creation of SuperTracy has been successful and contributed to the maturity of using generative AI solutions in PostNL, shedding light to its possibilities and business value, all done with the least amount of resources. This undermines the sometimes implicit assumption that adapting to generative AI technologies are expensive or out of reach.

\subsection{Future Work}

Future work depends on the scope, being the MVP or for scaling up through deployment. The MVP worked well, but could be improved. Future improvements to MVP of SuperTracy are refining the system’s ability to identify and communicate only the most relevant information, ensuring clarity and precision in the narratives. This can be done by removing auto generated default events in cases for events that can always be neglected. If events are interesting in some cases, a knowledge graph can be designed, which allows inference to understand the difference. Also the evaluation could be more extensive on diverse aspect. Human evaluation or Automatic evaluation methods could be used for aspects like fluency or multilingual performance. The additional aspects of the model could also be further evaluated, such as the individual performance of the logistic event prediction based on the T5 model. 

The broader scope can be the deployment of SuperTracy and further specifying its use-case. In the case of deployment, it is important to pay significant attention to data privacy and security, preventing oversharing sensitive information with the AI system, which are common challenges when leveraging LLMs in business.\cite{arman2023exploring}. This requires awareness of designing and using AI solutions in an enterprise. Next to the secure deployment, maintaining and increasing the data quality of the used data sets to fine-tune the LLM models are very important. In the evaluation some data quality issues have already been pointed out which deserve further attention. After all, the execution of this MVP has shown that the used methodologies provide value for the use-case. These same methods, technologies and approach can be used for other use-cases too.

\newpage
\appendix
\section*{Appendices}
\addcontentsline{toc}{section}{Appendices}

\section{Illustrations of various applications of SuperTracy, including user queries and system responses.}
\textit{*The illustrations include certain business-related information pertaining to PostNL, which has been redacted at the request of the company. }
\begin{figure}
    \centering
    \includegraphics[width=1\linewidth]{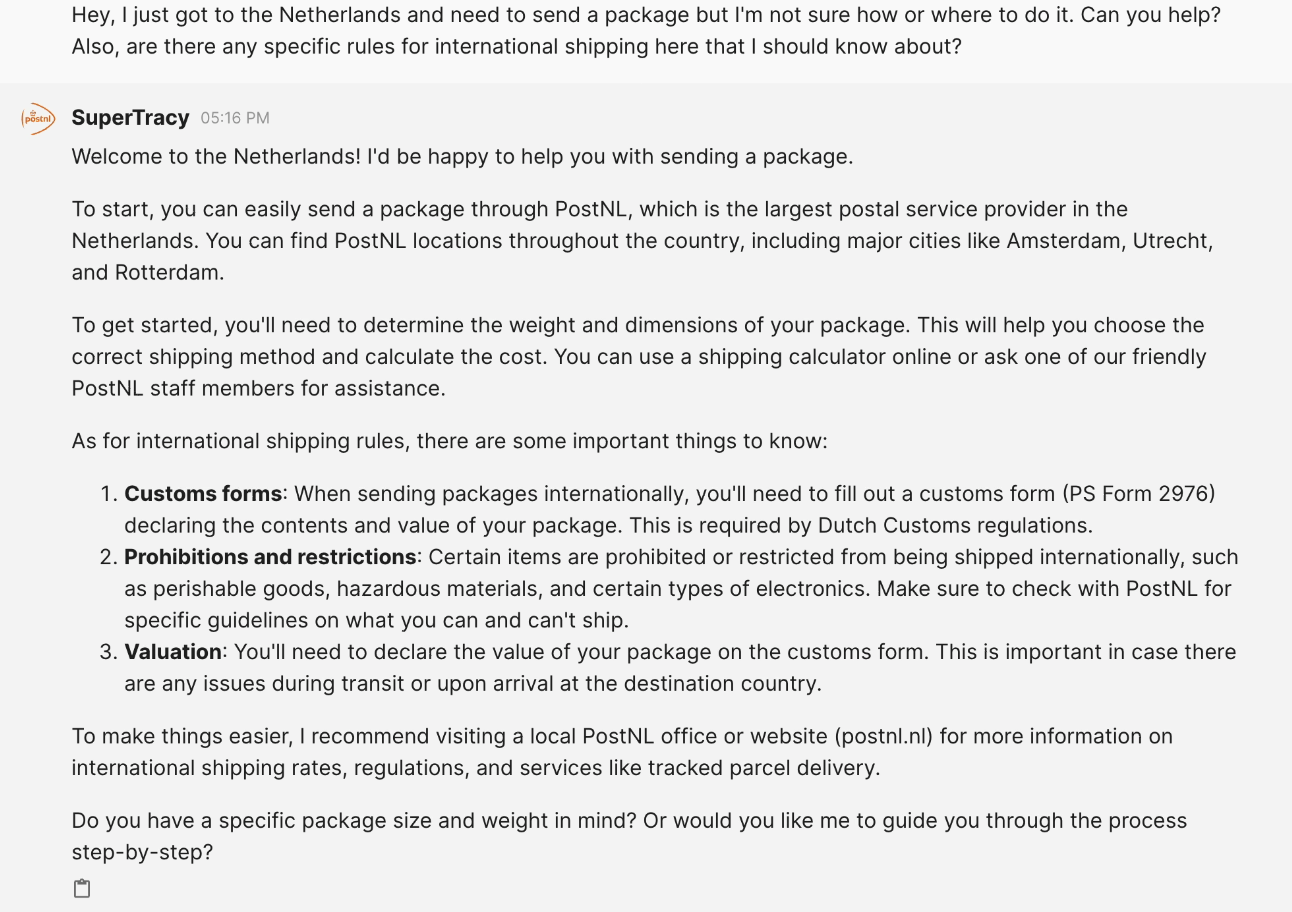}
    \caption{\textbf{Reception Agent Demo:}
Helping newcomers to the Netherlands send parcels back home }
    \label{fig:enter-label}
\end{figure}

\begin{figure}
    \centering
    \includegraphics[width=1\linewidth]{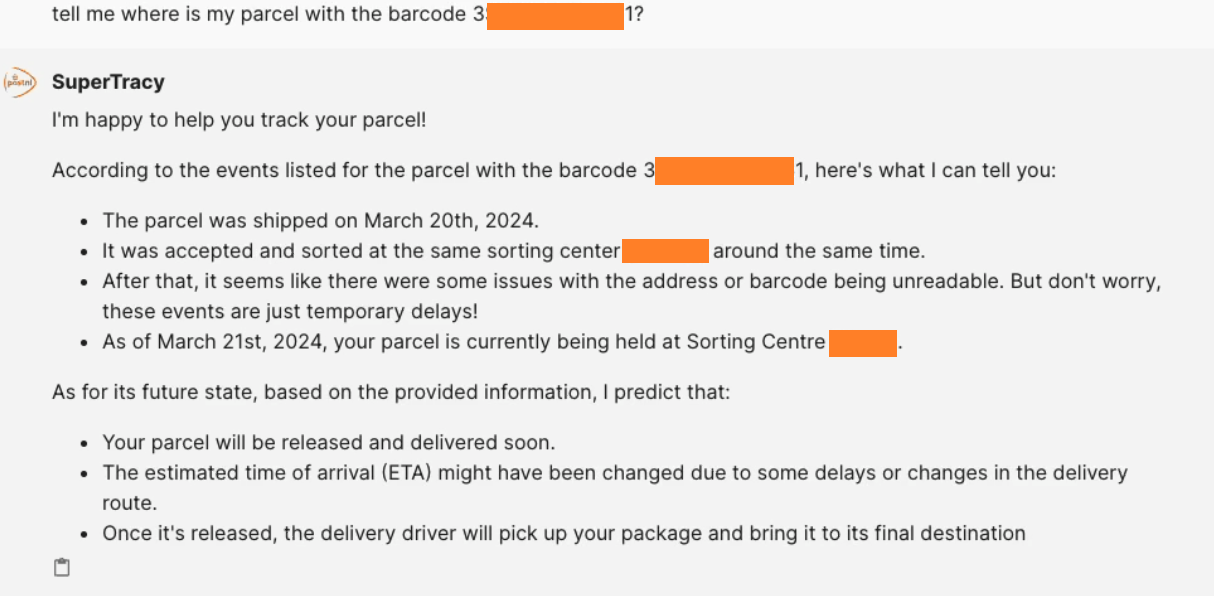}
    \caption{\textbf{Parcel Agent: }Predicting the next likely events of an undelivered parcel\textbf{\textit{\textbf{ }}  }}
    \label{fig:enter-label}
\end{figure}
\begin{figure}
    \centering
    \includegraphics[width=1\linewidth]{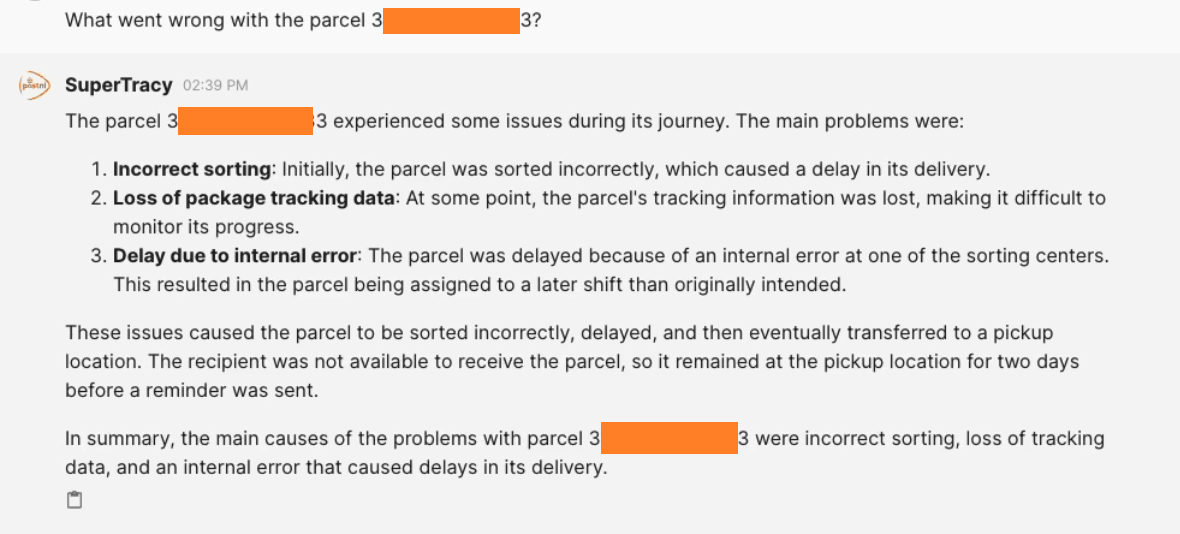}
    \caption{\textbf{Parcel Agent: } Recognizing an Unhappy Flow of a parcel, with feedback on Data Quality \textbf{  }}
    \label{fig:enter-label}
\end{figure}

\begin{figure}
    \centering
    \includegraphics[width=1\linewidth]{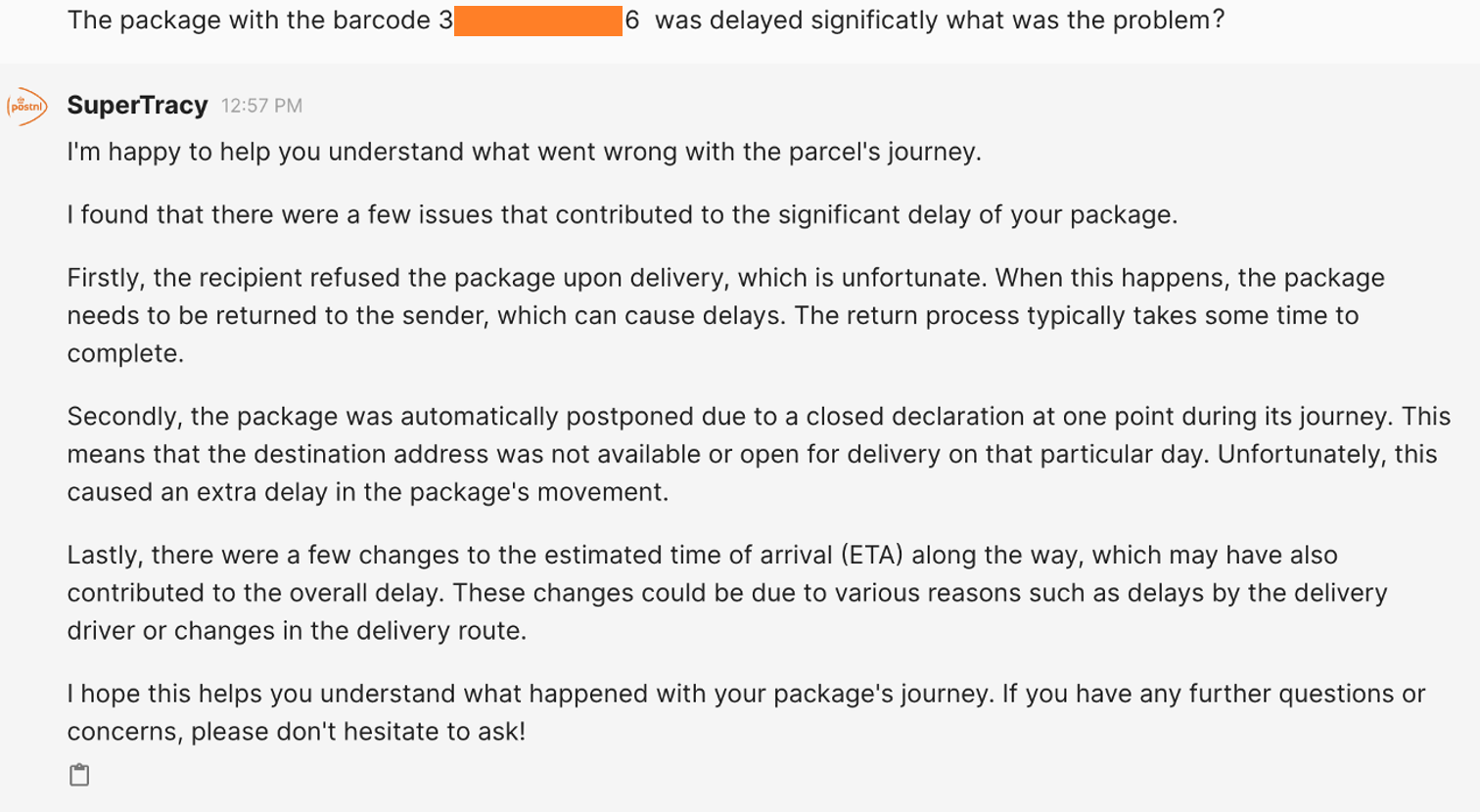}
    \caption{\textbf{Parcel Agent: }Analyzing a package that has been through a rough and unfortunate journey }
    \label{fig:enter-label}
\end{figure}
\begin{figure}
    \centering
    \includegraphics[width=1\linewidth]{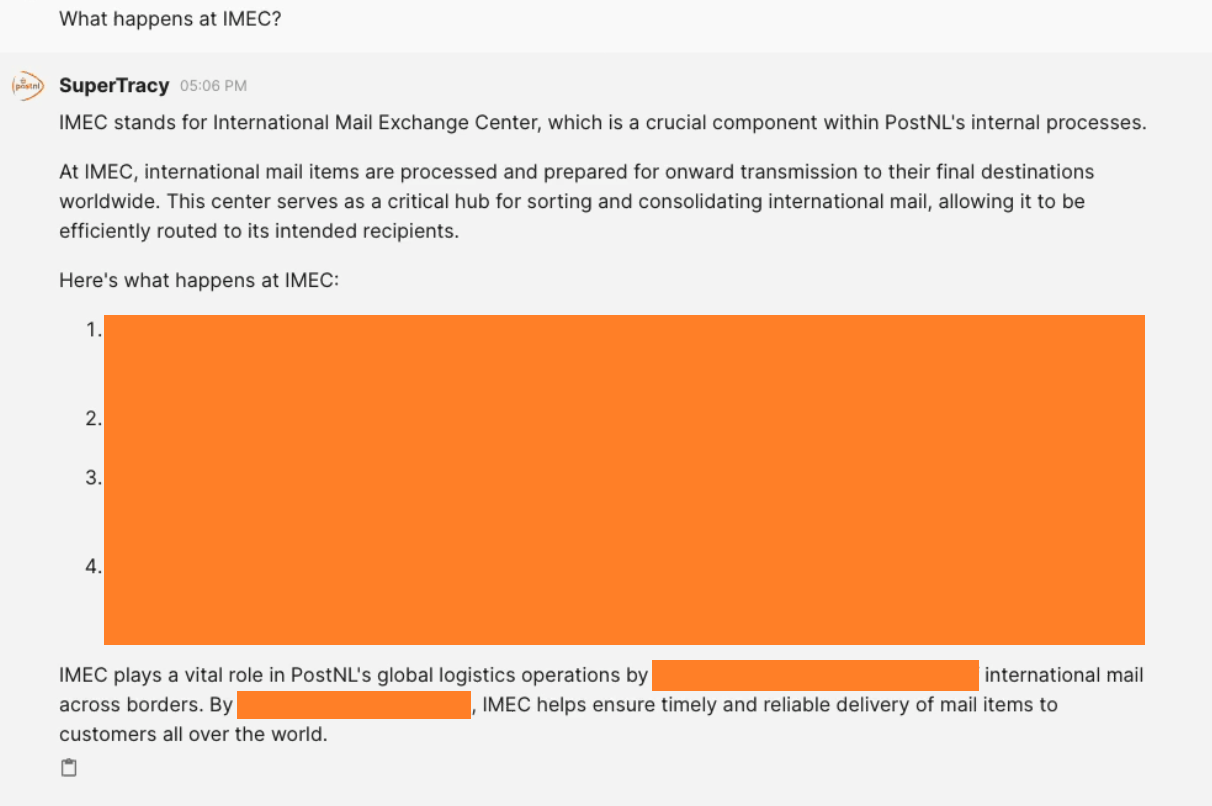}
    \caption{\textbf{Knowledge Expert Agent: }Asking about a PostNL internal specific process }
    \label{fig:enter-label}
\end{figure}

\begin{figure}
    \centering
    \includegraphics[width=1\linewidth]{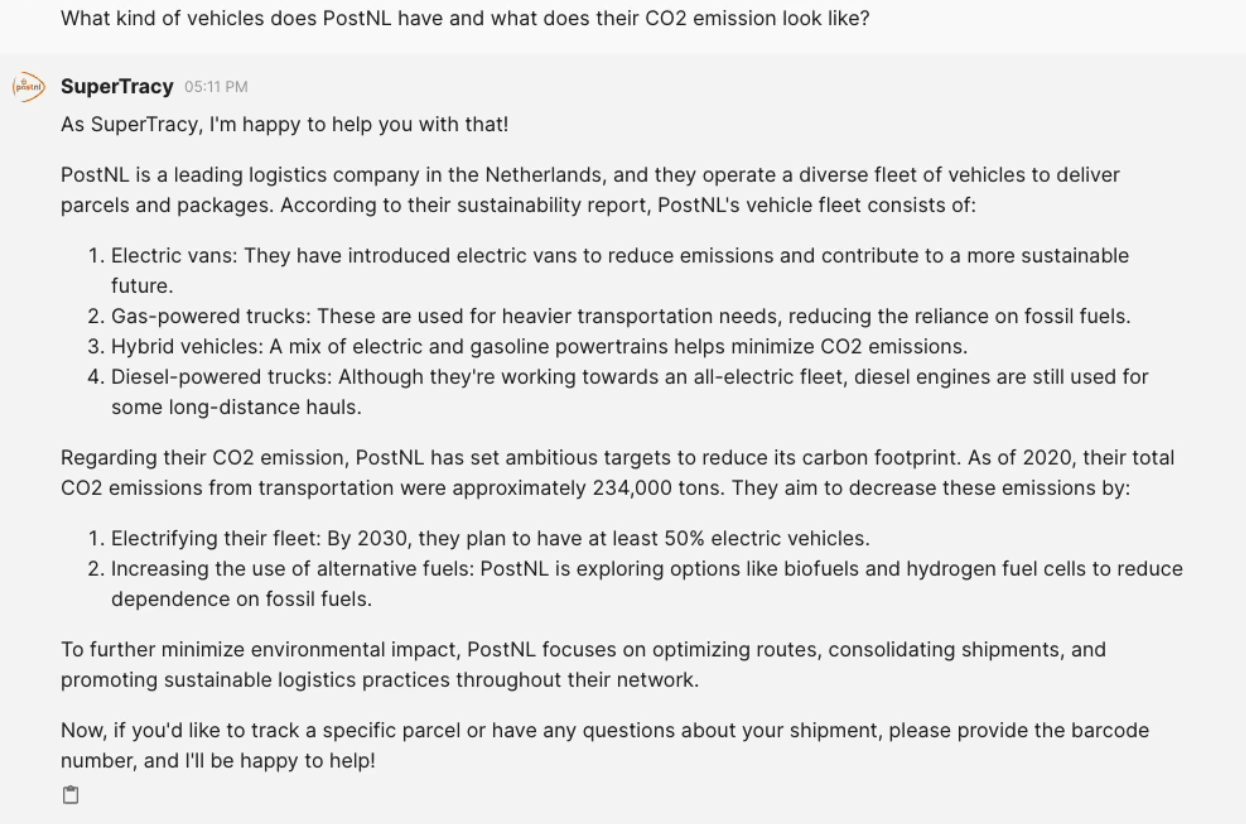}
    \caption{\textbf{Knowledge Expert Agent: }A general question on CO2 emissions of PostNL vehicles. }
    \label{fig:enter-label}
\end{figure}

\begin{figure}
    \centering
    \includegraphics[width=1\linewidth]{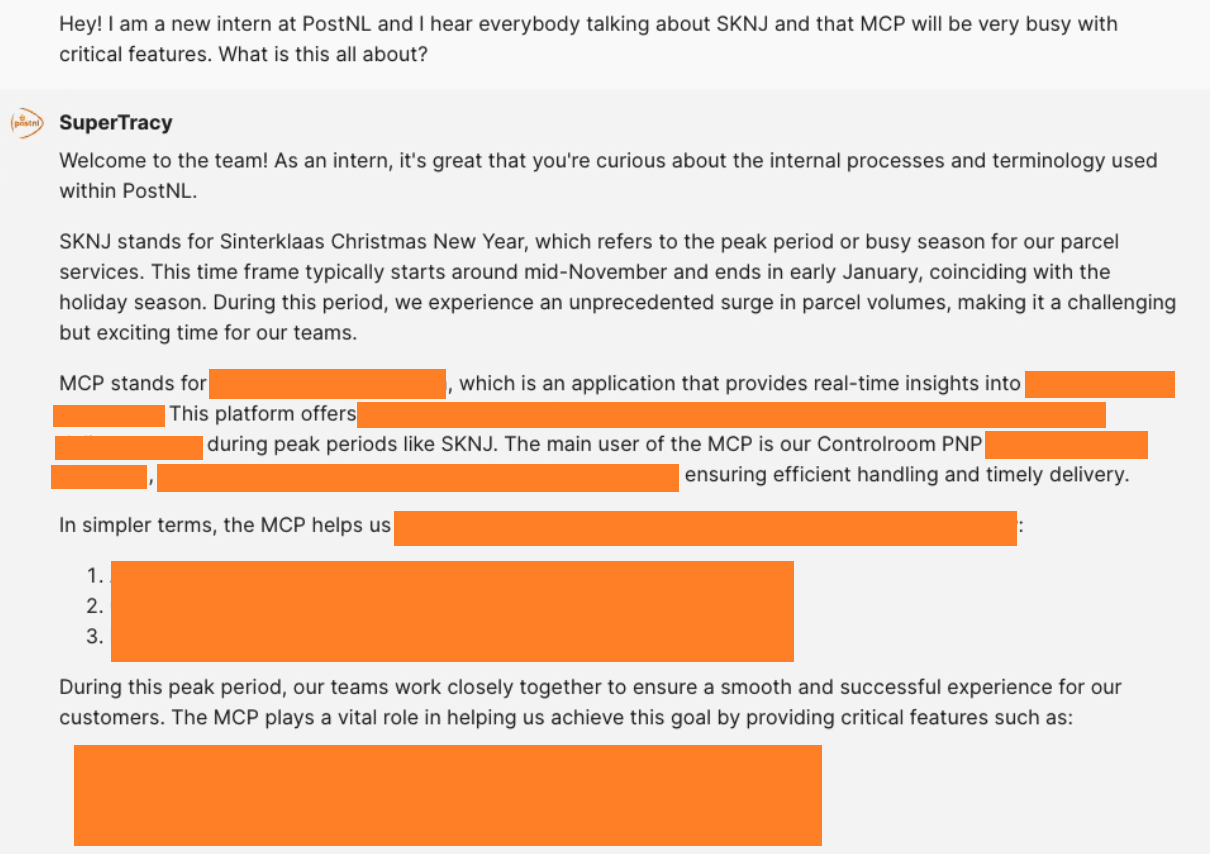}
    \caption{\textbf{Knowledge Expert Agent: }Helping a new intern with understanding PostNL abbreviations\textbf{ }\textbf{ }}
    \label{fig:enter-label}
\end{figure}

\begin{figure}
    \centering
    \includegraphics[width=1\linewidth]{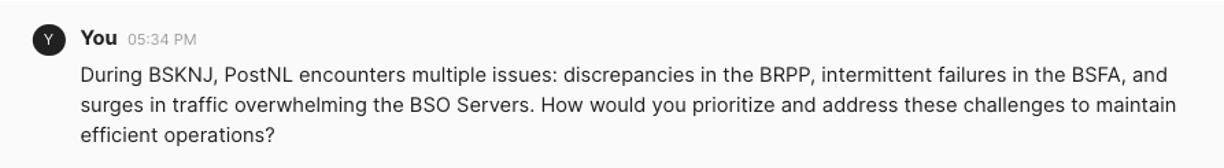}
    \caption{\textbf{Knowledge Expert Agent:} }Asking an advice on a PostNl specific complex situation (including PostNL abbreviations)\textbf{ }\textbf{ }
\end{figure}
\begin{figure}
    \centering
    \includegraphics[width=1\linewidth]{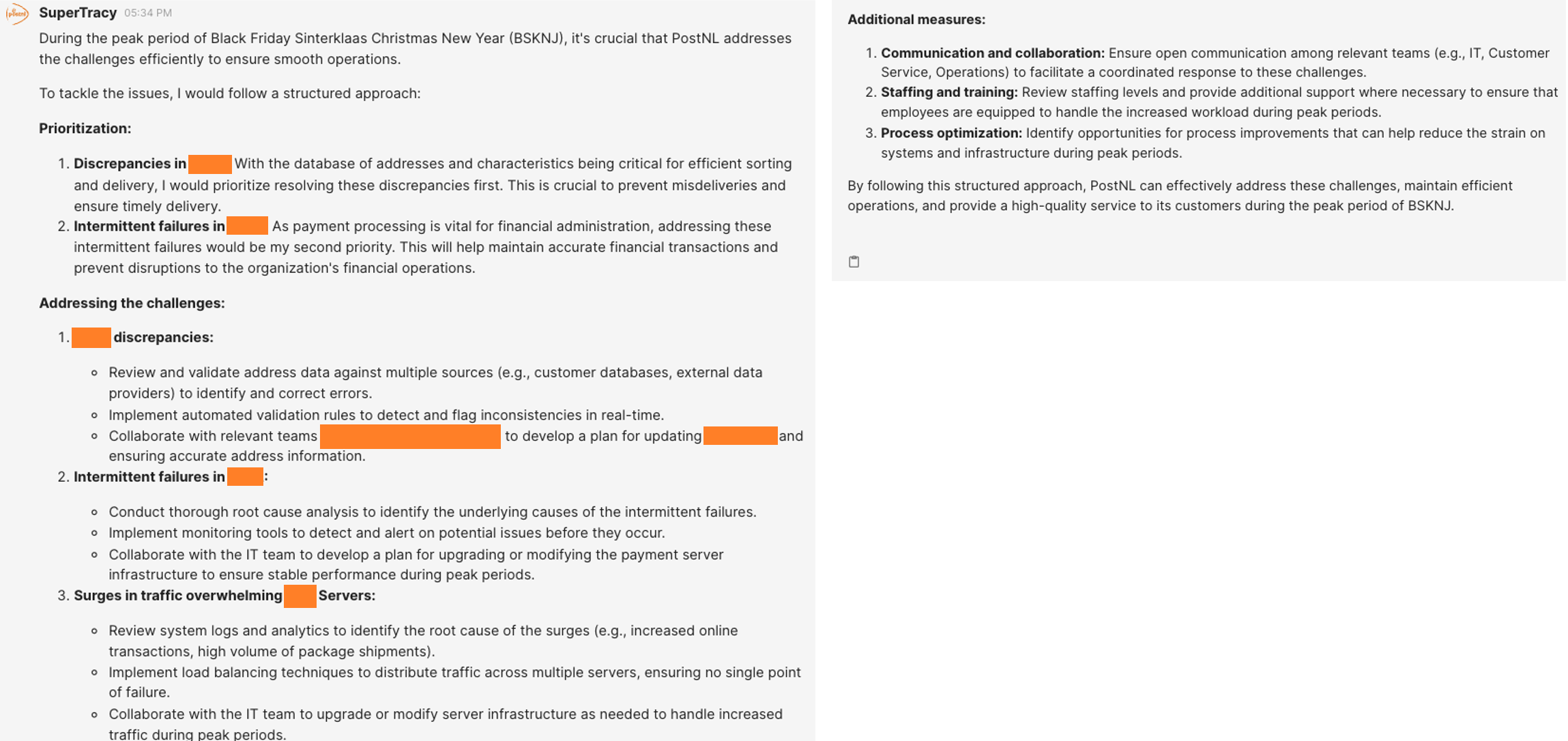}
    \caption{Respond to question in Figure 11}
    \label{fig:enter-label}
\end{figure}

\end{document}